# The Third Ambition:
# Artificial Intelligence and the Science of Human Behavior

W. Russell Neuman
New York University
wrn210@nyu.edu

Chad Coleman
New York University
cjc652@nyu.edu

## Abstract

Contemporary artificial intelligence research has been organized around two dominant ambitions: productivity, which treats AI systems as tools for accelerating work and economic output, and alignment, which focuses on ensuring that increasingly capable systems behave safely and in accordance with human values. This paper articulates and develops a third, emerging ambition: the use of large language models (LLMs) as scientific instruments for studying human behavior, culture, and moral reasoning. Trained on unprecedented volumes of human-produced text, LLMs encode large-scale regularities in how people argue, justify, narrate, and negotiate norms across social domains. We argue that these models can be understood as *condensates* of human symbolic behavior—compressed, generative representations that render patterns of collective discourse computationally accessible.

The paper situates this third ambition within long-standing traditions of computational social science, content analysis, survey research, and comparative-historical inquiry, while clarifying the epistemic limits of treating model output as evidence. We distinguish between base models and fine-tuned systems, showing how alignment interventions can systematically reshape or obscure the cultural regularities learned during pretraining, and we identify instruct-only and modular adaptation regimes as pragmatic compromises for behavioral research. We review emerging methodological approaches—including prompt-based experiments, synthetic population sampling, comparative-historical modeling, and ablation studies—and show how each maps onto familiar social-scientific designs while operating at unprecedented scale.

> Rather than proposing LLMs as substitutes for human subjects or causal explanation, we frame them as observational infrastructure analogous to large-scale datasets in genomics or astronomy. Properly constrained and triangulated, LLM-based analyses can reveal discursive priors, norm tradeoffs, framing sensitivities, and historical variation that complement traditional experiments, surveys, and ethnography. The third ambition thus expands the empirical field of the social and behavioral sciences by making the statistical structure of collective human expression newly observable, while demanding methodological discipline to avoid overinterpretation.

Since its modern resurgence, artificial intelligence research has been guided by two dominant ambitions. The first is productivity -- using AI systems to accelerate work, automate tasks, and increase economic efficiency. The second is alignment -- ensuring that increasingly capable systems behave safely, predictably, and in accordance with human values. Together, these ambitions have structured most contemporary debates about AI's promise and risks.

And now a third ambition is emerging -- less explicit, but potentially just as transformative. This ambition treats artificial intelligence not only as a tool for action or an object of control, but as a scientific instrument for studying human behavior itself. In this view, large-scale generative AI models become experimental mirrors of the cognitive, cultural, and moral patterns embedded in human societies.

Ambition One -- Productivity

The productivity ambition frames AI as a computational tool that can "do things." This notion draws on the seminal definition of artificial intelligence articulated by John McCarthy in his proposal for a seminal workshop study in the 1950s:

> "The study is to proceed on the basis of the conjecture that every aspect of learning or any other feature of intelligence can in principle be so precisely described that a machine can be made to simulate it." (McCarthy et al., 1955)

This definition explicitly treats intelligence as something that can be operationalized, formalized, and executed by a machine — i.e., intelligence as something that does things. The first six decades of AI research, typically characterized as the symbolic AI paradigm, followed a model of intelligence as the manipulation of hand-coded symbols and rules, requiring researchers to specify in advance how reasoning should proceed (McCarthy 1959; Newell & Simon 1976). Results were often incomplete and disappointing and critics were skeptical of its potential (Minsky & Papert 1969; Dreyfus 1992). Over the last decade, however, AI research shifted to a self-learning connectionist model primarily based on neural network architectures which reverses the original logic -- rather than encoding knowledge directly, models learn internal representations of large-scale data through optimization and feedback (Rumelhart et al. 1986; LeCun et al. 2015; Sutton 2019). Remarkable breakthroughs in systems that translate languages, generate code, classify images, optimize logistics, and produce text at scales previously unimaginable led to new enthusiasm about AI as an engine of economic growth. In this framing, success is measured by speed, accuracy, cost reduction, and substitution or augmentation of human labor (Autor et al. 2001; Brynjolfsson & McAfee 2014). This ambition has been remarkably successful and has attracted massive capital investment (Gil & Perrault 2025). AI increasingly functions as cognitive infrastructure—embedded in search engines, recommendation systems, workplace software, and creative tools. Much like earlier general-purpose technologies such as electricity or computing, AI is reshaping productivity not through single applications but through pervasive integration (Helpman 1998).

Ambition Two -- Alignment

The alignment ambition arose in response to the rapid growth of AI capabilities, as systems became more autonomous and generative and concerns about bias, deception, misuse, and loss of human control moved from speculative to operational (Bostrom 2014; Russell 2019). These concerns were not merely theoretical but were institutionalized in the founding of new organizations such as OpenAI, which initially framed its mission around the safe development of artificial general intelligence even as commercial pressures pushed the organization toward large-scale deployment. Tensions between model scaling and safety research later contributed to the formation of Anthropic as a spin-off explicitly oriented toward alignment-first development and constitutional constraints on model behavior. Alignment research itself has focused on embedding human norms, values, and constraints into AI systems, most notably through techniques such as reinforcement learning from human feedback (Christiano et al. 2017). By 2023, these anxieties had entered the public sphere through calls for a temporary pause in advanced AI development, reflecting a belief that governance and alignment mechanisms were lagging behind technical progress (Future of Life Institute 2023). The debate was further escalated by arguments from such activists as Eliezer Yudkowsky, who framed misaligned artificial general intelligence not as a manageable engineering risk but as a potential existential threat to humanity, thereby pushing alignment discourse beyond questions of safety and ethics toward species-level survival (Yudkowsky 2023). Teams of researchers at MIT, Cambridge, Carnegie Mellon, and UC Berkeley as well as OpenAI, Anthropic, Google and AI2 are pursuing alignment research (Christian 2020; Slattery et al. 2024; Gil & Perrault 2025). Here the guiding question is no longer what AI can do, but how it should behave. Alignment treats AI systems as entities whose outputs can have social, ethical, and political consequences, even if they lack intentions or consciousness.

A Third Ambition -- Understanding Human Culture and Behavior

Beyond productivity and alignment lies a third ambition: using artificial intelligence as a tool for social and behavioral science. Large language models are trained on massive corpora of human-produced text—books, news, legal opinions, online discussions, moral arguments, religious writings, and everyday conversations—now commonly measured in the tens of trillions of tokens (Kaplan et al. 2020; Heikkilä & Arnett 2024). In absorbing these materials, models internalize not only linguistic structure, but patterns of reasoning, norm enforcement, moral disagreement, emotional expression, and cultural variation as a cultural and behavioral condensate at a scale that no prior empirical method has approached.

Never before have researchers had access to a single representational system trained on such a vast, heterogeneous archive of human symbolic behavior—an accessible and responsive archive that reflects how people explain themselves, justify actions, argue about values, negotiate social roles, and encode collective memory across domains and institutions. While these models do not "understand" in a characteristic human sense, they function as compressed statistical condensates of how humans have thought, spoken, and reasoned in public and semi-public contexts. As such, they offer a new observational lens on human psychology and culture, analogous in ambition—if not in method—to large-scale genomic databases or astronomical sky surveys.

This creates a novel scientific opportunity. For the first time, researchers can systematically probe a system trained on an enormous cross-section of human symbolic output and ask structured, repeatable questions about what it has learned. How do humans reason about fairness versus loyalty? Under what conditions do people privilege outcomes over rules? When do moral judgments shift in response to authority, group membership, uncertainty, or perceived threat? By varying prompts, contexts, and framing, researchers can observe consistent response patterns that reflect aggregate regularities in human discourse rather than idiosyncratic individual responses.

Crucially, AI models do not provide causal explanations in the way formal psychological or sociological theories do. They struggle to distinguish explanation from correlation, and intention from imitation. However, they can reveal the statistical structure of human judgment, belief, and norm articulation at extraordinary scale and resolution, offering a powerful empirical complement to surveys, experiments, ethnography, and historical analysis (Argyle et al. 2023; Horton 2023). In this role, large language models function as high-dimensional mirrors of collective human cognition.

Several scholars and commentators have already gestured toward this possibility. Work on large language models as "simulators" of human populations has shown that model responses often reproduce known psychological and political regularities found in survey data, suggesting their potential as synthetic—but informative—samples of human judgment (Argyle et al. 2023). Others have argued that foundation models constitute a new kind of cultural artifact: trained not on curated theories but on the sedimented outputs of human society itself (Bommasani et al. 2021; Manning et al. 2020). Journalistic and philosophical treatments have similarly suggested that large language models provide a distorted but revealing portrait of contemporary culture, capturing its obsessions, conflicts, blind spots, and moral anxieties (Christian 2020).

At the same time, any such use must acknowledge important limitations. The corpus of human text on which contemporary models are trained is incomplete and uneven. It overrepresents Western, industrialized, and English-language sources; privileges literate, online, and institutionally mediated forms of expression; and underrepresents oral traditions, marginalized communities, and non-digitized cultures (Bender et al. 2021). As a result, the "human thought" encoded in these models is neither universal nor neutral. It is a historically contingent, technologically mediated sample—valuable precisely because it reveals which voices dominate global discourse in the digital age, which norms are most frequently articulated, and which perspectives are systematically absent.

Seen in this light, the third ambition of artificial intelligence is not to replace social science, but to augment existing methods with a new kind of empirical object: a computationally tractable condensation of collective human expression. Properly interpreted and carefully constrained, analysis of these trillions of tokens offers an unusual opportunity to study human psychology and culture not as abstract ideals, but as they are enacted—messily, unevenly, and at scale—in the language humans use to make sense of their world.

What Does Generative LLM Output Represent?

John McCarthy's famous mission statement quoted above posits "intelligence can in principle be so precisely described that a machine can be made to simulate it." McCarthy was bold in his thinking but his choice of words was cautious. He used the general term "intelligence" rather than human intelligence. But most readers, then and now, we suspect, would interpret the word as "human intelligence," or "human-level intelligence." The inevitable evocation is the question – AI as smart as or smarter than humans, and perhaps even somehow human? The literature that preceded McCarthy made that clear – case in point – the Turing Test (Turing 1950). And the current and dramatically expanding literature on Artificial General Intelligence (AGI) following McCarthy builds on that tradition of questioning the dangers of and capacities of machines to be human-like (Bostrom 2014; Russell 2019; Yudkowski 2023). As a result, reviewers of behavioral and cultural research based in whole or in part on generative output raise objections to the effect that generative output isn't authentically human. So how can we best describe what generative output can be said to represent?

We are in novel territory here, so some fresh terminology may be useful. Author and critic Ted Chiang used the compression metaphor noted above in his *New Yorker* article "ChatGPT Is a Blurry JPEG of the Web" (2023). Drawing on the distinction between lossy and lossless compression he notes that there is a big gap between the 15 trillion training words (tokens) and the resultant neural network matrix that represents ChatGPT's generative engine. The model isn't the training data, it is a reflection of it. How best can we characterize this epistemologically?

To be clear in our vocabulary we need distinguish the model and the model's output. The model itself is a Learned Distribution of Human Symbolic Behavior. That is a mouthful, so we use the term Condensate as a shorthand. This is a conditional probability structure learned from data, that cannot be directly observed. What we have access to is the Generative Output (GO), a realized sample of text, images, choices, rationales derived from the model in response to prompts. So, following this perspective, the Condensate represents a reflection of a limited sample of human symbolic behavior and the GO represents the sampled outputs of explicit prompts corresponding roughly to the outputs of traditional social science interviews, surveys and experiments. This vocabulary makes three things clear. 1) The model has been mathematically trained. Even modestly different mathematical equations based on the same exact data could result in distinctly different models. 2) The training is based on a specific dataset. Although most model designers have been strategically vague about the data used, we understand the data is a highly constrained and unrepresentative sample of human symbolic behavior, broadly defined. 3) Because the model is distributive, it is likely to generate different Generative Output to even slight variations in prompting text.

This vocabulary invites the researcher to compare condensates to each other and to other assessments of human behavior. If differently designed models, and models trained on distinctly different datasets generate the same response, that is a finding. If models generate the same response to a systematically varied set of prompts, that is a finding. If the Generative Outputs diverge in systematic ways, the question becomes what does that tell us about the Condensates?

We will explore examples of these comparative dynamics in the section below on Methodological Innovations.

The Fine-Tuning Issue

The modern productivity and alignment ambitions do not utilize raw self-taught "base-model" condensates. To make the base models more responsive to prompts and more aligned with the values and preferences of their designers, they are heavily modified in a layered process dubbed fine-tuning (Balavadhani et al 2024). Fine-tuning typically further trains models on curated question–answer examples and structured human feedback. Question–answer tuning improves task responsiveness and coherence, while reinforcement learning from human feedback (RLHF) further shapes outputs by rewarding clarity, helpfulness, politeness, and topical focus. The raw training data includes material that reflects historical prejudice, offensive language, and harmful stereotypes. Typically, this involves filtering and suppressing outputs associated with hate speech, explicit violence, criminal instruction, harassment, or discriminatory language; discouraging racist, sexist, or otherwise demeaning stereotypes; and constraining the model from generating instructions for self-harm, fraud, or other forms of wrongdoing, even when such patterns are present in the raw training data (Christiano et al. 2017; Ouyang et al. 2022). These interventions explicitly encode contemporary human values and legal norms into system behavior, often by training models to refuse certain requests, redirect conversations, or adopt socially acceptable framing.

Fine-tuning represents a unique challenge to the Third Ambition as it can distort historical or factual representations, as illustrated by widely publicized cases in which content filters or diversity constraints produced implausible outputs—such as historically inaccurate depictions associated with image-generation systems at Google (Morrone 2024). Such cases underscore that fine-tuning functions as a normative control layer rather than a neutral correction, requiring continuous adjustment to balance factual accuracy, social harm reduction, and user trust. In this way, modern AI systems operationalize productivity and alignment as sequential stages in a single pipeline: general intelligence is learned first, and social acceptability is imposed afterward. It could be said that fine-tuning steps us back to the rule-based, semantic, hand-coded paradigm of AI as a result of our concern about misbehavior as a result of what these models learned about how humans actually communicate and behave. We will return to these challenges below.

The Incomplete Sample Issue

As noted above, we have incomplete information about the provenance of the tens of trillions of tokens typically scraped from the web to train most modern LLMs (Liu et al. 2024; Bommasani et al. 2021). It is highly likely, however, that most models are trained on basically the same fundamental database with minor exceptions as it represents what is currently accessible at appropriate scale including the Common Crawl, Wikipedia, multiple book archives, and multiple social media archives (Wiggins 2025).

Concern persists that this common core represents a biased and incomplete proxy for human culture as web text over-represents networked, literate, platform-active populations (Bender et al. 2021; Blodgett et al. 2020; Resnik 2025). As a result many analysts recommend rebalancing strategies including identifying representational skews (language, geography, dialect, and topic domains) and corrective strategies involving targeted supplementation and weighting of the web corpora.

Case studies of prominent webtext pipelines show that "cleaning" and filtering can systematically remove particular varieties of language while retaining other institutional sources in unexpected quantities—making the dataset's social worldview an artifact of pipeline choices, not merely of the web itself (Dodge et al. 2021). This has motivated calls for dataset "datasheets" that specify composition, collection, and intended uses, and for dataset-development practices that include validation of representativeness relative to the target population or phenomenon (Gebru et al. 2018; Paullada et al. 2021). Taken together, these proposals make rebalancing a three-part program: 1) document and audit training data, 2) deliberately supplement and weight to address known skews, and 3) report model behaviors and limits in a way that supports accountable use (Bender et al. 2021; Resnik 2025).

Complementing the Existing Large Language Models with New Data

Researchers may be drawn to "subsample" the extant massive aggregate database on which the LLMs are pretrained by constraining the models through prompting to utilize only training data from different languages, geographic areas, historical periods or cultural traditions thus permitting comparative analysis. For example, one could develop a prompting routine that requires the model to draw only on Hindi language training data in responding to a particular ethical dilemma (Agarwal et al. 2024). In our research we have found that models do respond with agility to such prompting constraints and provide systematically varied responses based on the required constraints (Coleman et al. 2026). But because the pretraining process is holistic, the "subsampled" data is not stored in addressably separate parts of the model and the subsampling process is inherently inferential and indirect.

Fortunately, there are a variety of technical methodologies that strengthen comparative research designs by introducing new, structured data into large language models without modifying the base model parameters or sacrificing their general-purpose generative capacities. Among the most prominent are parameter-efficient fine-tuning approaches, which add small, modular components to pretrained models while leaving the original weights frozen. Low-Rank Adaptation (LoRA), for example, injects trainable low-rank matrices into selected layers of a pretrained transformer, allowing models to acquire domain-, language-, or culture-specific competencies with orders of magnitude fewer parameters than full fine-tuning (Hu et al. 2021). Because the base model remains unchanged, LoRA adapters can be added, removed, or swapped, enabling controlled comparisons across differently conditioned variants that share a common representational core.

This modularity has made LoRA and related approaches particularly attractive for comparative and counterfactual analyses. Researchers have used LoRA adapters to specialize models for specific languages, registers, or domains, effectively constructing parallel "views" of the same underlying model trained on distinct subsets of data (Lester et al. 2021; Dettmers et al. 2023). In contrast to prompt-based constraints—which rely on the model's internal inference over entangled representations—adapter-based methods create explicit, addressable differences in how information is processed, while preserving the shared inductive biases learned during large-scale pretraining. This makes it possible to attribute observed differences in model behavior more directly to differences in data exposure rather than to stochastic variation in generation or prompt interpretation.

Closely related approaches extend this logic beyond fine-tuning to memory- and retrieval-based augmentation, in which models are coupled to external datastores or auxiliary modules that can be selectively constrained at inference time. Retrieval-augmented architectures and nearest-neighbor language models condition generation on explicitly defined corpora—such as texts from a particular historical period, geographic region, or cultural tradition—without altering the pretrained model's parameters (Khandelwal et al. 2020; Borgeaud et al. 2021). Together, these methods form a continuum of strategies for comparative analysis: from indirect inferential subsampling via prompting, to modular parameter-efficient adaptation, to explicit external memory constraints. Each progressively increases experimental control over the cultural and historical inputs brought to bear in model responses, while retaining the generative power of large-scale pretrained language models.

A New Observational Tool for the Social Sciences?

If the social science community approached governments and foundations for $4B to build a model of human thought and communication with which to test their theories, one might expect a skeptical response. In contrast, when the physics community made the case for that level of funding for the Large Hadron Collider (LHC) to test their theories they were notably successful. Ironically and incidentally something in that range of investment has resulted in the publicly accessible Large Language Models which although designed explicitly for ambitions one and two, offer great promise for creative new methodologies and a new shared resource for the social sciences.

What makes the LHC analogy instructive is not simply the scale of investment, but its illustration of the way in which new scientific instruments can historically transform what scientific questions can be asked and answered. Many of the most consequential advances in science have not arisen from new theories alone, but from new ways of observing phenomena that were previously inaccessible (Kuhn 1962; Hacking 1981). The microscope did not invent cell theory, but by rendering microbial life visible it made cellular explanations unavoidable, reshaping biology by expanding the empirical domain to which theory had to answer (Latour 1987). Similarly, the telescope did not resolve cosmological debates by itself, but it expanded the observable universe in ways that forced revisions to long-standing assumptions about celestial order and motion. In each case, instruments did not replace theory; they reorganized the relationship between theory and evidence by vastly enlarging the observable field.

A similar dynamic can be seen in the rise of large-scale observational datasets across the sciences. In economics, the emergence of administrative data, high-frequency transaction records, and platform-scale behavioral logs transformed the field from one dominated by small samples and stylized models into one increasingly grounded in population-level behavioral traces (Einav & Levin 2014). These data infrastructures did not eliminate the need for causal inference or theory-driven explanation, but they enabled the discovery of regularities—such as persistent heterogeneity, rare but consequential tail events, and path-dependent dynamics—that were difficult or impossible to detect with traditional survey or experimental methods alone. Genomics offers an even clearer example: the Human Genome Project did not explain biological function by itself, but it created a shared empirical substrate that reoriented entire disciplines, enabling new forms of hypothesis generation, comparison, and validation across biology and medicine (Leonelli 2016).

Large language models can be understood as part of this same instrumental lineage. They do not provide direct access to "human thought," just as telescopes do not provide direct access to cosmological truth or genomic databases do not explain phenotypes on their own. Instead, they constitute a new class of observational infrastructure: systems trained on unprecedented volumes of human symbolic output that render patterns of discourse, reasoning, and normative judgment computationally accessible. What is novel is not that human language has become data—social scientists have long analyzed texts—but that a single, unified representational system now encodes an enormous cross-section of that symbolic behavior in a form that can be systematically probed, perturbed, and compared By presenting models with carefully designed prompts—moral dilemmas, political conflicts, ambiguous narratives, or social trade-offs—researchers can observe consistent response patterns that reflect the underlying distribution of human discourse on which the models were trained. Prompting becomes a form of computational experiment.

The Antecedents of the Third Ambition

Long before machine learning, scholars treated texts not merely as carriers of meaning but as behavioral artifacts: systematic residues of social life that could be collected, coded, and analyzed to reveal latent structures of belief, power, and coordination. The Third Ambition represents a natural extension of this research tradition under conditions of unusual scale, heterogeneity, and generativity.

The first of these antecedents lies in early traditions of text as social signal, developed most clearly in mid-twentieth-century content analysis. Harold Lasswell's wartime studies of propaganda and political communication framed text as observable evidence of political intent and social influence, crystallized in his canonical formulation—"who says what, to whom, with what effect" (Lasswell 1948). Bernard Berelson subsequently systematized content analysis as a reproducible social-scientific method, emphasizing reliability, coding schemes, and inferential validity (Berelson 1952). This tradition reached an early computational apogee with Philip Stone's General Inquirer, which translated psychological and moral categories into machine-readable dictionaries, enabling large-scale automated analysis of meaning (Stone et al. 1966). The key conceptual move here was epistemic: texts could be treated as measurable outputs of social systems, suitable for aggregation, comparison, and theory testing.

A second tradition conceptualized text as cultural archive, emphasizing scale and historical depth rather than immediate signaling. Projects such as the Human Relations Area Files transformed ethnographic texts into structured, comparable data, enabling cross-cultural inference from narrative sources (Murdock & White 1969; Murdock 1971). Decades later, large digital corpora—most prominently Google Books and the Ngram Viewer—extended this logic dramatically, allowing scholars to trace diachronic shifts in language use, norms, and collective attention across centuries (Michel et al. 2011). Although methodologically crude by contemporary standards, these efforts established a powerful premise: culture leaves statistical fingerprints in large bodies of text, and those fingerprints can be interrogated quantitatively to study social change.

The third strand emerged in the early 2000s with the maturation of text as latent measurement. Advances in political methodology and computational linguistics reframed text analysis as a problem of estimating unobserved positions—ideology, framing, agendas—from observed word distributions (Grimmer et al. 2022). Models such as Wordscore, Wordfish, and later structural topic models demonstrated that texts could be used to place actors, institutions, and time periods within shared latent spaces, enabling comparison across cases and over time (Blei et al. 2003). This work was central to the consolidation of computational social science as a field, showing that textual data could support formal inference, cumulative theory building, and hypothesis testing rather than merely descriptive analysis.

Finally, a fourth tradition reconceptualized text as cultural residue, emerging from the analysis of digital platforms, search logs, and social media. Internet search data, in particular, were reframed as a form of revealed preference—linguistic behavior produced outside survey contexts and often under conditions of anonymity (Stephens-Davidowitz 2017). This work blurred the boundary between communication and behavior, treating text as a byproduct of action rather than a self-conscious expression of belief. The methodological lesson was that large-scale textual traces could reveal patterns systematically inaccessible through self-report alone.

Seen against this backdrop, the Third Ambition appears as a natural evolution of these converging traditions. Large language models can be understood as compressed, generative representations of the very textual distributions that earlier scholars painstakingly collected and analyzed: propaganda corpora, ethnographies, books, political speeches, and digital traces. What is new is not the ambition to learn about human culture from text, but the epistemic stance that treats LLMs themselves as analyzable cultural artifacts—synthetic yet grounded in real distributions—capable of being probed, compared, and situated within long-standing traditions of computational inquiry into social life.

Skepticism About LLMs as a Tool for Social and Behavioral Research

Despite growing interest in large language models as potential instruments for studying human culture and behavior, the existing literature reflects deep and persistent ambivalence about whether, and under what conditions, such an enterprise is epistemically justified. Across psychology, sociology, communication, ethics, and ethnographic research, scholars have raised concerns that treating LLMs as windows into human cognition and collective culture risks conflating linguistic fluency with psychological validity, statistical regularities with lived experience, and convenience with inference (Bender et al. 2021; Rossi 2024; Schroeder et al. 2025).

A central line of skepticism centers on the problem of psychological validity and the risk of mistaking linguistic mimicry for cognitive competence (Grizzard 2025; Schroeder et al. 2025). Critics argue that large language models generate fluent responses by statistically reproducing patterns in human speech rather than by engaging in genuine judgment, reasoning, or moral cognition. This might be characterized as the Chinese Room Critique, drawing on John Searle's famous thought experiment positing an individual with no understanding of Chinese who could appear to understand by closely following a set of instructions about manipulating Chinese symbols (Searle 1980). The concern has been crystallized in the AI era by the characterization of generative AI as a "stochastic parrot" that echos human language without understanding its meaning or grounding (Bender et al. 2021). Much of this critique is framed by the question of whether synthetic personas created by AI can "substitute" for human ones (Wang et al. 2024; Boelaert et al. 2025). The correlation of survey response means between human and synthetic personas is remarkably high – a correlation of .95, for example, in one meta-review (Dillion et al. 2023). James Bisbee and colleagues (2024) note this pattern of high similarity and ask "But does this imply that we can replace human samples altogether?" When framed that way, they raise an appropriately cautionary note. But in our view, such a frame precludes the notion that these generative models can supplement and expand traditional rather than simply replace research methods. Bisbee and colleagues go on to make the point that in multivariate analyses some variations between human and synthetic responses are evident and that estimates of variances in synthetic data tend to be smaller. They insist that since the internal workings of these LLMs are, in effect, a black box, in their terms – "an algorithm with unknown properties," this line of research is fraught with risk. True, but that might be said of the complexity of logic and emotion that intertwine in human survey responses. In our thinking, if generative and human responses differ in systematic ways – that in itself is an engaging question to explore.

Another strand of skepticism concerns the limited representativeness of the training data. Critics stress that the textual corpora used to train contemporary models are neither comprehensive nor representative of humanity writ large. Instead, they overrepresent Western, English-language, digitally mediated discourse, embedding the priorities, norms, and asymmetries of global information infrastructures (Bender et al. 2021; Bommasani et al. 2021). From this view, LLMs may reveal important features of dominant or institutionalized culture, but they cannot be treated as neutral summaries of "human culture" without reproducing the very exclusions and power dynamics scholars seek to analyze.

Another element of this literature notes conceptual slippage—the tendency to anthropomorphize models or treat them as psychological entities rather than technical artifacts. Some authors explicitly reject framing LLMs as possessing beliefs, intentions, well-being, or reflexivity, warning that such language unthinkingly imports human-centered concepts into systems that lack embodiment, development, or agency (Jowsey et al. 2025; Mims 2025). Even work exploring model "introspection" or internal consistency often cautions that these behaviors should be interpreted as properties of trained representations, not evidence of self-awareness or cognitive states (Bender et al. 2024; Shojaee et al. 2025).

Importantly, the literature is not uniformly dismissive. Several authors articulate a more qualified or instrumental stance, acknowledging both the risks and the potential value of LLMs for cultural and behavioral research if their limitations are made explicit. Abdurahman et al. (2024) describe the use of LLMs in psychology as a domain of "perils and opportunities," arguing that models can be informative when used for hypothesis generation, scale exploration, or methodological triangulation rather than inference about individuals. Similarly, Buttrick (2024) frames LLMs as compression algorithms for human culture, a metaphor that captures both their analytical promise and their distortions. From this perspective, the question is not whether LLMs *are* human culture, but what kinds of cultural regularities become visible through their compression. Works by Rus and Mone (2024) and broader ethical surveys of AI scholarship similarly underscore the tension between interpretive opportunity and epistemic overreach, particularly when models are used outside their original design intentions (Windsor 2021).

Taken together, this literature does not consistently reject the use of LLMs for understanding human culture and behavior outright, but it strongly cautions against naive or substitutionist approaches. The dominant skeptical position holds that LLMs cannot replace human subjects, ethnography, experiments, or theory-driven explanation. However, a narrower consensus is emerging around a more limited claim: that LLMs may serve as instruments for probing the structure of human symbolic behavior, provided they are treated as artifacts of specific data regimes, governance choices, and alignment interventions rather than as models of human minds. The challenge, then, is not whether LLMs can be used in the social and behavioral sciences, but whether they can be used with sufficient methodological discipline to avoid mistaking cultural compression for cultural understanding.

Perhaps the best strategy for moving forward with the third ambition is a cautious experimentalism – defining the comparability of LLMs and human speech and behavior as a testable experimental question developing new methodological designs and carefully comparing and contrasting their findings with traditional ones. Scholars should treat LLMs as computational instruments for observing patterns in human symbolic behavior, rather than as agents, theories of mind, or substitutes for human subjects.

Comparing Human Learning and Machine Learning

At its core, skepticism toward the use of large language models in the social and behavioral sciences rests on a well-founded premise: machine learning and human learning are not the same kind of process, and any comparison between them risks obscuring fundamental differences. Human learning is embodied, affectively charged, developmentally staged, and socially scaffolded. It unfolds through lived experience—sensorimotor interaction, emotional regulation, and reinforcement through biologically salient rewards and punishments in a dynamic physical and social environment. Human learning begins long before language, as infants acquire causal knowledge by acting on the world—grasping objects, observing consequences, and updating expectations through prediction error—rather than by passively observing symbolic descriptions (Piaget 1954; Gibson 1979).

From the earliest months of life, infants integrate perception, action, and affect. They learn object permanence, agency, and intentionality through repeated cycles of exploration and feedback, guided by innate attentional biases and dense social scaffolding (Baillargeon 2004; Spelke & Kinzler 2007). Crucially, this learning is grounded in multimodal input—visual, auditory, tactile, proprioceptive—and regulated by motivational systems tied to survival, attachment, and social belonging. Language acquisition itself is built atop this prelinguistic substrate: words are mapped onto objects, actions, and goals that are already meaningful through shared attention and embodied interaction (Tomasello 1999). Nothing in contemporary large language models reproduces this developmental trajectory, motivational architecture, or embodied grounding.

For this reason, we do not treat large language models as developmental analogues of human learners, nor as models of individual cognition (Piantadosi & Hill 2022). The comparison we draw operates at a narrower and more abstract level: the level of how structured competence can emerge from prolonged exposure to patterned environments under iterative feedback, rather than from the mechanisms, motivations, or experiences through which that competence is acquired.

At this abstract level, a parallel becomes analytically informative. In pretraining, large language models expose initially unstructured parameter matrices to vast streams of symbolic input, optimizing predictions through gradient-based error correction (Brown et al. 2020; Kaplan et al. 2020). In the earliest stages of training—much as in the earliest stages of human learning—there is little internal structure on which to base accurate prediction, and error rates are correspondingly high. Through repeated cycles of exposure and optimization, models gradually acquire structured regularities that encode relationships among entities, actions, causes, and social expectations. These regularities are shaped not by embodiment or affect, but by inductive patterns embedded in model architecture, objective functions, and training data (Sutton 2019; Bommasani et al. 2021; Khatri et al. 2025). Where human learners interact with a physical and social world, large language models interact exclusively with a symbolic one (Futrell & Mahowald 2025; Gandhi et al. 2023).

Seen in this light, the relevance of large language models for social and behavioral research does not depend on their resemblance to human learners as organisms. Instead, it rests on a different, more limited claim: LLMs internalize the accumulated symbolic traces of human social learning at scale, including the narratives, norms, justifications, and disagreements through which cultures transmit knowledge. In this respect, they exaggerate a real feature of human cognition—our capacity to learn from testimony, imitation, and cultural inheritance rather than from first-person experience alone. As development proceeds, human learners increasingly rely on social learning, narrative, and linguistic instruction to acquire knowledge far beyond what direct interaction would permit (Gergely & Csibra 2006; Henrich 2016).

The Base Model and the Fine-Tuning Problem

The promise of large language models for understanding human culture and behavior rests on their exposure to unprecedented volumes of human-produced text. Contemporary frontier models are trained on trillions of tokens drawn from web data, books, news, forums, and institutional documents, capturing large-scale regularities in how humans reason, justify, argue, narrate, and normatively regulate social life (Penedo et al. 2023; Talmor et al. 2019). However, the very systems made available to researchers are not neutral interfaces to this "raw" cultural corpus. Instead, they are the product of extensive post-training interventions—filtering, fine-tuning, reinforcement learning, and policy-driven constraint—that substantially distort access to the underlying statistical structure of the training data (Bender et al. 2021; Abdurahman et al. 2024).

This creates what could be termed the fine-tuning problem for social and behavioral research. While base models encode broad patterns of human symbolic behavior, deployed models are deliberately reshaped to comply with safety standards, politeness norms, legal constraints, and institutional value commitments. Alignment techniques—including reinforcement learning from human feedback, constitutional prompting, and values-targeted datasets—systematically suppress, reframe, or redirect outputs associated with taboo topics, moral conflict, political extremism, hate speech, or social deviance (Solaiman and Dennison 2021; Anthropic 2023).

The difficulty is compounded by the practical inaccessibility of base models. While base or minimally trained models are, in principle, the closest approximation to unfiltered cultural data, interacting with them directly is often infeasible. Base models are typically unstable, incoherent, unsafe, or restricted from public access, making them unsuitable for systematic probing or reproducible research. As a result, researchers are forced to operate on models that already embed multiple layers of post-hoc constraint, obscuring the boundary between patterns learned from human text and norms imposed by institutions (Linegar et al. 2023; Magee et al. 2023).

Recent mechanistic and cognitive studies underscore how consequential these interventions can be. Fine-tuning has been shown to alter internal representations, decision pathways, and reasoning strategies in ways that are not merely superficial but structurally transformative (Jain et al. 2023; Binz et al. 2025). Alignment procedures can increase conformity, politeness, and refusal behavior while simultaneously reducing variance, suppressing minority perspectives, and flattening culturally specific modes of reasoning (Yin et al. 2023; Wang et al. 2024). From the standpoint of social science, this means that heavily aligned models risk reflecting idealized institutional morality rather than the contested, heterogeneous moral landscapes present in the underlying data.

At the same time, emerging work in cognitive modeling and linguistics suggests that pretraining itself captures meaningful structure related to human cognition and social reasoning. Foundation models trained on large-scale text have been shown to predict aspects of human behavioral data, align with neural signals during language processing, and reproduce patterns associated with social inference and theory of mind (Gandhi et al. 2023; Gao et al. 2025; Futrell and Mahowald 2025). These findings reinforce the view that the representational substrate learned during pretraining is analytically valuable—but only if researchers can distinguish it from subsequent normative overlays.

Given these constraints, it may be that instruct-only fine-tuning represents the most viable methodological compromise for social and behavioral research. Instruct-only tuning aims to improve coherence, responsiveness, and task compliance without introducing heavy normative filtering, values-based reward shaping, or extensive refusal policies (Balavadhani et al. 2024). Compared to reinforcement-based alignment or constitutional training, instruct-only models preserve greater access to the distributional structure of the underlying text while remaining usable, interpretable, and safe enough for systematic study. Although instruct-tuned models are still shaped by human preferences, they minimize the imposition of explicit moral or political constraints that would otherwise confound cultural analysis.

This position aligns with a growing recognition that alignment is not a binary property but a spectrum of interventions with distinct epistemic consequences (Anthropic 2025; Senthilkumar et al. 2024). For deployment, strong alignment may be essential. For social science, it is an obstacle. As several authors note, alignment mechanisms can obscure precisely those tensions, ambiguities, and norm violations that are most informative for understanding human behavior (Abdurahman et al. 2024; Rossi 2024). Treating aligned models as transparent mirrors of culture risks mistaking curated civility for genuine consensus.

In short, the fine-tuning problem forces a methodological choice. Either researchers accept heavily filtered outputs that reflect institutional norms more than cultural reality, or they adopt carefully constrained interaction regimes—such as instruct-only tuning—that preserve access to the representational residue of large-scale human discourse while maintaining analytical control. Our approach embraces the latter strategy, not because it restores access to "raw" human culture in any pure sense, but because it offers the clearest available window onto the symbolic structures learned during pretraining. For the third ambition of artificial intelligence—using LLMs as tools for understanding human culture and behavior—this distinction is not ancillary; it is foundational (Shah et al. 2026).

Methodological Innovations

The Third Ambition aligns with evolving traditions in computational social science that analyze large-scale cultural traces to identify regularities, boundaries, and structures in human communication (Salganik 2017; Windsor 2021; Shah et al. 2015). Rather than introducing wholly new methods, large language models recombine and extend established social-science approaches by altering scale, accessibility, and the nature of the empirical object itself. The methodological contribution of LLM-based research is therefore best understood in terms of the potentially new types of findings it enables, the theoretical assumptions it places under pressure, the range of disciplines it connects, and the hybrid qualitative–quantitative methods it makes feasible.

Computational Social Science emphasizing text-as-data has accumulated a variety of labels in various partially overlapping social science traditions over the past 75 years including: quantitative text analysis, content analysis, frame analysis, discourse analysis, corpus linguistics, and computational linguistics. In each of these traditions the methodology drew on a unique textual data set and a particular analytic tool such as a topic model or content analytic codebook. The critically important development now is the emergence of the modern neural-net-based Large Language Models following the introduction of ChatGPT in the fall of 2022.

But two significant elements have changed. There is a singular and extremely large corpus (virtually identical for most modern LLMs) and the breadth of semantic and statistical tools for accessing it have expanded dramatically. This is a paradigm shift of potentially Kuhnian proportions. If one wants to compare the texts of Shakespeare and Moliere, the propaganda of the Germans and the Americans during WWII, the recent policy proposals of Republicans and Democrats it is all available for analysis through multiple theoretical lenses. One can systematically manipulate prompts to simulate different experimental conditions to simulate experiments or surveys of different populations of interest. We will review a small sample of the evolving work to illustrate (See also, Appendix A).

Computational Experiments – An Extension of the Experimental Tradition

Aher et al. (2023), for example, explicitly frame LLMs as experimental subjects, reproducing well-known psychological and economic experiments by varying prompt instructions and contextual framing while holding task structure constant. By manipulating features such as payoff framing, social roles, and cooperative incentives, they show that LLM outputs shift in predictable ways across experimental conditions, often mirroring—but sometimes diverging from—human response patterns. Importantly, when generated results diverge from characteristic human behavior, they need not be defined as invalid or without relevance. On the contrary, it raises the fresh research question of what in the comparative character of machine learning and human learning would lead to such systematic difference (Yax et al. 2024; Niu et al. 2024). Neuman et al. (2025) examines the ethical reasoning of eight prominent generative large language models in response to classic moral dilemmas such as the Trolley Problem and the Heinz Dilemma. The results largely mirror parallel human-based studies with the models consistently weighing consequentialist analytics more heavily than deontological ones.

Related work has focused on how discursive framing and authority cues systematically shape model behavior. Germani and Spitale (2025) demonstrate that identical informational content produces measurably different outputs depending on whether it is attributed to authoritative institutions, expert sources, or anonymous individuals. By varying only the source framing in otherwise identical prompts, they identify stable representational sensitivities to epistemic authority. Similarly, Cheung et al. (2025) apply canonical decision-theoretic manipulations— gain versus loss framing, certainty versus uncertainty, and reference point shifts—to show that large language models exhibit amplified versions of well-documented cognitive biases under controlled prompt variation. In both cases, prompt perturbation operates as a computational analogue of experimental manipulation, revealing causal dependencies within the model's learned representations.

A third line of work adapts cultural transmission and moral psychology paradigms to the prompt domain. Takemoto (2024) conducts large-scale "moral machine" experiments on language models by systematically varying the structure of moral dilemmas. The resulting shifts in moral judgments closely parallel classic vignette studies in moral psychology, illustrating how prompt-based perturbations reflect structured normative patterns. Complementing this, Acerbi and Stubbersfield (2023) embed language models in iterated transmission chains, repeatedly perturbing prompts across generations to study how narratives distort over time. Taken together, these studies establish experimental prompt perturbation as a robust tool for probing representational structure in large language models, closely aligned with—but methodologically distinct from—traditional experimental research on human subjects.

Synthetic Personas – An Extension of the Survey Tradition

Synthetic population sampling parallels survey research by eliciting structured responses under systematically varied conditions, such as persona constraints or hypothetical social positions. Classical survey methods aim to estimate attitudes, beliefs, or preferences across populations through standardized instruments (Converse 1964; Groves et al. 2009). In contrast, LLM-based sampling is proposed here to map discursive priors—the range and distribution of responses that are culturally plausible within the model's learned representation.

Qu & Wang (2024) explore the policy tradeoffs between economic and environmental goals in simulated populations of American, Japanese, South African and Brazilian populations. Park et al. (2024) present a novel agent architecture that simulates the attitudes and behaviors of 1,052 real individuals--applying large language models to qualitative interviews about their lives, then measuring how these agents respond to classic attitudinal and behavioral measures. Aher et al (2023) study how perceptions of fairness and obedience behavior vary by gender. Neuman et al. (2025b) investigates the political temperaments of seven prominent LLMs using a set of established political ideology scales. They find strong and consistent prioritization of liberal-leaning values, particularly care and fairness, across most measures.

Computational Comparative-Historical Analysis

A prominent example of the promise of LLMs for comparative historical research is offered by Varnum et al. (2024), who argue for the construction of "historical language models" trained on time-bounded corpora drawn from specific historical periods. They propose training or fine-tuning models on texts from distinct eras—such as classical antiquity, the medieval period, or early modern Europe—and then probing those models with standardized psychological or moral questions. Their central claim is that the models encode statistically recoverable regularities in the values, concerns, and explanatory styles that dominated written discourse at different moments in time.

Related work extends this logic by linking LLM outputs directly to longitudinal survey data and value change. Liu et al. (2025) compare model-generated responses to multiple waves of the World Values Survey and explicitly ask which historical periods' attitudes contemporary LLMs most closely resemble. Coleman et al. (2026) contrast the ethical logic of five world religions by training separate models independently on 100 million tokens drawn from the online Sacred Text Archive. Cassotti & Tahmasebi (2025) use fine-tuned LLMs to generate sense-specific historical word usage, constructing diachronic datasets that capture how meanings evolve across periods. Hauser et al. (2024) introduce HiST-LLM, a benchmark grounded in historical knowledge from the Seshat Global History Databank, helping establish shared resources for evaluating how well language models encode temporally specific information.

Computational Cognitive Neuroscience

The phrase computational cognitive neuroscience may strike some as a contradiction in terms. The numeric mechanics of the AI matrices and the electrochemistry of the human brain are distinctly different phenomena. But comparing and contrasting the internal workings of the two can be revealing. The traditions of alignment research and AI explainability have led to new models of research to make more accessible the internal workings of the famous AI "black box." Chris Olah leads a research team at Anthropic which produced the widely cited "Mapping the Mind of a Large Language Model" (2024) which uses a case study of how an Anthropic neural network internally represents meaningful concepts – referred to as features – as directions in their activation spaces. They use the term monosemanticity as a goal in AI interpretability research – locating identifiable features or units such as "neurons" or "activations" within a neural network which can be said to represent a single specific concept. Recent work in large scale interpretability demonstrates that large language models contain thousands of neurons corresponding to stable, human-interpretable features that can be automatically described and empirically validated, reinforcing the view that these models encode structured regularities of human discourse rather than opaque statistical noise (Choi et al., 2024; Dasdan et al. 2026).

Computational Ablation Studies – An Extension of Computational Social Science

Ablation studies are adapted from machine learning as a means of identifying functional dependencies by selectively removing or suppressing components of a system (LeCun et al. 2015). In the context of studying human culture and behavior through LLMs, ablation is repurposed as a form of computational counterfactual cultural analysis.

We consider several forms of ablation: (1) data ablation, in which specific categories of training text (e.g., legal discourse, religious writings, narrative fiction) are reduced or excluded; (2) alignment ablation, in which fine-tuning constraints are relaxed to observe masked representations; and (3) representational ablation, where identifiable internal pathways associated with particular concepts are weakened or removed. These approaches parallel lesion studies in neuroscience and institutional counterfactuals in historical sociology, allowing researchers to ask which cultural supports are necessary for particular forms of reasoning or norm articulation to appear (Hassabis et al. 2017; Sutton 2019). Ablation does not reveal causal origins of human behavior, but it does illuminate symbolic dependencies—the cultural infrastructures on which certain patterns of discourse rely.

Validation, Triangulation, and Limitations

Consistent with best practices in social science methodology, findings derived from LLM-based methods should be evaluated through explicit forms of triangulation rather than treated as stand-alone evidence (Denzin 1978; Salganik 2018). We do not interpret LLM outputs as indicators of individual cognition, population prevalence, or causal mechanism. Instead, validation targets a narrower class of claims: the structure, directionality, and contextual sensitivity of symbolic responses as they appear in large-scale human discourse. In this framework, successful validation does not require exact correspondence with human data, but convergence at the level of relative patterns—such as tradeoffs, prioritizations, or framing effects—documented in existing empirical literatures. LLM outputs are not evidence of individual cognition or population prevalence. Instead, they are analyzed as structured reflections of historically contingent, power-weighted, and institutionally mediated human discourse (Bender et al. 2021; Bommasani et al. 2021).

Triangulation operates along several concrete dimensions. First, convergent validation compares LLM-derived patterns to well-established findings from surveys, experiments, and comparative research; agreement in direction or ordering, rather than magnitude, constitutes support, while systematic reversals without theoretical explanation count against validity. Second, boundary and discriminant checks assess whether LLMs differ from human-based data in predictable ways in domains where embodiment, emotion, or lived experience are known to be decisive; such patterned limitations are informative, whereas unrealistic uniformity or context-insensitivity signals overreach. Third, robustness tests probe sensitivity to prompt framing, sampling parameters, and model choice; findings that persist across reasonable perturbations are treated as stable, while results that collapse under minor variation are discounted.

Importantly, this approach is falsifiable. LLM-based methods would be undermined if they failed to produce reproducible patterns across repeated sampling, converged toward generic or institutionally sanitized responses regardless of prompt structure, or yielded differences that could not be traced to identifiable features of training data or fine-tuning regimes. For these reasons, LLM outputs are best understood as hypothesis-generating observations. Their epistemic value lies in expanding the observable field of human symbolic behavior and revealing large-scale regularities and tensions that can then be tested, refined, or rejected through established human-subject, historical, and ethnographic methods. As with other observational sciences, credibility accrues not from any single result, but from cumulative replication, comparative analysis, and disciplined triangulation.

To be clear, validation does not require exact correspondence with human data; discrepancies are expected and are treated as hypotheses-generating signals that prompt targeted comparative analysis rather than as grounds for rejection.

The Third Ambition — Next Steps

The third ambition introduces a new perspective on how generative AI models might be able to contribute to human productivity and self-understanding. Large language models were not built to explain human behavior, yet by virtue of their training on unprecedented volumes of human-produced text, they encode regularities in reasoning, norm enforcement, moral disagreement, and cultural framing that no prior method could capture at comparable scale. Treated appropriately, these models offer a new cultural and behavioral condensate for the social and behavioral sciences—one that complements experiments, surveys, ethnography, and historical analysis without replacing them. The challenge is not to anthropomorphize these systems or mistake fluency for understanding, but to recognize that compression itself can be analytically revealing.

The next steps for this Third Ambition are therefore methodological in character. Progress depends on developing disciplined research designs that make explicit what is being probed, what is being held constant, and what kinds of inference are and are not warranted. Prompt perturbation, synthetic population sampling, representational content analysis, and ablation studies provide concrete analogues to classical social-science methods, enabling systematic exploration of symbolic dependencies, normative boundaries, and cultural variation. At the same time, the fine-tuning problem underscores the need for transparency about model regimes: heavily aligned systems reflect institutional values as much as cultural ones, while instruct-only tuning offers a pragmatic compromise for research purposes. Validation through triangulation with human-subject data remains essential, not as a benchmark of success but as a safeguard against overinterpretation. The Third Ambition advances not by claiming privileged access to human nature, but by carefully delineating the conditions under which LLM outputs can be informative.

Finally, realizing this aspiration will require institutional recognition that large language models constitute a shared scientific resource, not merely a commercial product or governance challenge. Just as the Large Hadron Collider created common infrastructure for physics, frontier language models—however imperfect and contested—have already created a de facto observatory for human symbolic behavior. Whether this resource is used responsibly and productively by the social sciences will depend on norms of access, documentation, replication, and critique. The most promising posture is neither technological enthusiasm nor categorical skepticism, but cautious experimentalism: treating the comparability of LLM outputs and human discourse as an empirical question to be tested rather than assumed. If pursued in this spirit, the third ambition may not only deepen our understanding of artificial intelligence, but also sharpen our theories of human behavior itself.

References

Abdurahman, Suhaib, et al. (2024). "Perils and opportunities in using large language models in psychological research." PNAS Nexus 3(7).

Acerbi, Alberto and Joseph M. Stubbersfield (2023). "Large language models show human-like content biases in transmission chain experiments." Proceedings of the National Academy of Sciences 120(44): e2313790120.

Agarwal, Utkarsh, et al. (2024). "Ethical Reasoning and Moral Value Alignment of LLMs Depend on the Language we Prompt them in." ArXiv: arXiv:2404.18460.

Anthropic (2023). "Claude's Constitution." Anthropic.

Anthropic (2025). "System Prompts." anthropic.com.

Argyle, Lisa P., et al. (2023). "Out of One, Many: Using Language Models to Simulate Human Samples." Political Analysis 31(3): 337-351.

Autor, David H., et al. (2001). "The skill content of recent technological change: An empirical exploration." The Quarterly Journal of Economics 118(4): 1279-1333.

Baillargeon, Renée (2004). "Infants' physical world." Current Directions in Psychological Science 13(3): 89-94.

Balavadhani Parthasarathy, Venkatesh, et al. (2024). "The Ultimate Guide to Fine-Tuning LLMs from Basics to Breakthroughs: An Exhaustive Review of Technologies, Research, Best Practices, Applied Research Challenges and Opportunities." arXiv: arXiv:2408.13296.

Bender, Emily M., et al. (2021). "On the Dangers of Stochastic Parrots: Can Language Models Be Too Big?" Proceedings of the 2021 ACM Conference on Fairness, Accountability, and Transparency: 610–623.

Berelson, Bernard (1952). Content Analysis in Communications Research. New York, Free Press.

Berger, Peter L. and Thomas Luckmann (1966). The Social Construction of Reality: A Treatise in the Sociology of Knowledge. Garden City, NY, Anchor.

Biber, Douglas, et al. (1998). Corpus linguistics: investigating language structure and use. Cambridge: Cambridge University Press.

Binz, Marcel, et al. (2025). "A foundation model to predict and capture human cognition." Nature 644(8078): 1002-1009.

Bisbee, J., et al. (2024). "Synthetic Replacements for Human Survey Data? The Perils of Large Language Models." Political Analysis 32(4): 401-416.

Blei, David M. (2012). "Probabilistic Topic Models." Communications of the ACM 55(4): 77-84.

Blodgett, Su Lin, et al. (2020). "Language (Technology) is Power: A Critical Survey of "Bias" in NLP." Proceedings of the 58th Annual Meeting of the Association for Computational Linguistics: 5454-5476.

Boelaert, J., et al. (2025). "Machine Bias: How Do Generative Language Models Answer Opinion Polls?" Sociological Methods & Research.

Bommasani, Rishi, et al. (2021). "On the Opportunities and Risks of Foundation Models." arXiv: arXiv:2108.07258.

Borgeaud, Sebastian, et al. (2021) "Improving language models by retrieving from trillions of tokens." arXiv, arXiv:2112.04426 DOI: 10.48550/arXiv.2112.04426.

Bourdieu, Pierre (1991). Language & Symbolic Power. Harvard University Press.

Bostrom, Nick (2014). Superintelligence: paths, dangers, strategies. Oxford, UK, Oxford University Press.

Brown, Tom B., et al. (2020). "Language Models Are Few-Shot Learners." ArXiv.

Brynjolfsson, Erik and Andrew McAfee (2014). The second machine age: work, progress, and prosperity in a time of brilliant technologies. New York, W. W. Norton & Company.

Buttrick, Nicholas (2024). "Studying large language models as compression algorithms for human culture." Trends in Cognitive Sciences 28(3): 187-189.

Cassotti, Pierluigi and Nina Tahmasebi (2025). "Sense-specific Historical Word Usage Generation." Transactions of the Association for Computational Linguistics 13: 690-708.

Chaudhary, Yaqub (2024). "Islam and Artificial Intelligence". In The Cambridge Companion to Religion and Artificial Intelligence.

Cheng, Zerui, et al. (2025) "Benchmarking is Broken -- Don't Let AI be its Own Judge." arXiv:2510.07575

Cheung, Vanessa, et al. (2025). "Large language models show amplified cognitive biases in moral decision-making." Proceedings of the National Academy of Sciences 122(25): e2412015122.

Chiang, Ted (2023). "ChatGPT Is a Blurry JPEG of the Web." The New Yorker.

Choi, Dami, et al. (2024). "Scaling Automatic Neuron Description." Transluce(October 23).

Christian, Brian (2020). The Alignment Problem: Machine Learning and Human Values. New York, W. W. Norton & Company.

Christiano, Paul, et al. (2017) "Deep reinforcement learning from human preferences." arXiv:1706.03741

Coleman, Chad, et al. (2026). "Six Llamas: Using Large Language Models to Explore Comparative Ethics." New York University.

Converse, Philip (1964). "The Nature of Belief Systems in Mass Publics". Ideology and Discontent. David Apter. New York, Free Press: 206-261.

Dasdan, Ali, et al. (2026). "The Neuronal Logic of Ethical Choice " New York University.

Denzin, Norman K. (1978). The research act: A theoretical introduction to sociological methods. New York, McGraw-Hill.

Dettmers, Tim, et al. (2023) "QLoRA: Efficient Finetuning of Quantized LLMs." arXiv, arXiv:2305.14314

Dillion, Danica, et al. (2023). "Can AI language models replace human participants?" Trends in Cognitive Sciences 27(7): 597-600.

Dodge, Jesse, et al. (2021) "Documenting Large Webtext Corpora: A Case Study on the Colossal Clean Crawled Corpus." arXiv, arXiv:2104.08758 DOI: 10.48550/arXiv.2104.08758.

Dreyfus, Hubert L. (1992). What Computers Can't Do. Cambridge, MIT Press.

Einav, Liran and Jonathan Levin (2014) "The Data Revolution and Economic Analysis." Journal of Economic Perspectives. 3-26.

Fasching, Darrell et al. (2011). Comparative Religious Ethics. Wiley.

Foot, Phillipa (1967). "The Problem of Abortion and the Doctrine of the Double Effect." Oxford Review 5.

Future of Life Institute (2023) "Pause Giant AI Experiments: An Open Letter."

Gandhi, Kanishk, et al. (2023). "Understanding Social Reasoning in Language Models with Language Models." arXiv: arXiv:2306.15448.

Gao, Changjiang, et al. (2025). "Increasing alignment of large language models with language processing in the human brain." Nature Computational Science.

Gebru, Timnit, et al. (2018) "Datasheets for Datasets." arXiv, arXiv:1803.09010 DOI: 10.48550/arXiv.1803.09010.

Gergely, Gréta and Gergely Csibra (2006). "Sylvia's Recipe: The Role of Imitation and Pedagogy in the Transmission of Cultural Knowledge". Roots of Human Sociality: Culture, Cognition, and Human Interaction. N. J. Enfield and S. C. Levenson. Oxford, Berg**:** 229-255.

Gibson, James J. (1979). The ecological approach to visual perception. Boston, Houghton Mifflin.

Gil, Yolanda and Raymond Perrault (2025). AI Index Report 2025. Stanford, Stanford Institute for Human-Centered Artificial Intelligence.

Gould, Hannah and Keiko Nishimura (2024). "The Buddha in AI/Robotics". The Cambridge Companion to Religion and Artificial Intelligence. Beth Singer and Fraser Watts. (eds.) Cambridge, Cambridge University Press: 50-68.

Grimmer, Justin, et al. (2022). Text as Data: A New Framework for Machine Learning and the Social Sciences. . Princeton Princeton University Press.

Grimmer, Justin and Brandon M. Stewart (2013). "Text as Data: The Promise and Pitfalls of Automatic Content Analysis Methods for Political Texts." Political Analysis 21(3): 267-297.

Grizzard, M. (2025). "ChatGPT Does Not Replicate Human Moral Judgments: The Importance of Examining Metrics Beyond Correlation to Assess Agreement." Scientific Reports.

Groves, R. M., et al. (2009). Survey Methodology. New York, Wiley.

Gudorf, Christine E. (2013). Comparative religious ethics: everyday decisions for our everyday lives. Minneapolis, Fortress Press.

Hacking, Ian (1981). Scientific Revolutions. Oxford, Oxford University Press.

Hargittai, Eszter (2008). "The Digital Reproduction of Inequality". Social Stratification. David Grusky. Boulder, CO, Westview Press: 936-944.

Hassabis, Demis, et al. (2017). "AI and Neuroscience: A virtuous circle." Google Deep Mind.

Heikkilä, Melissa and Stephanie Arnett (2024). "This is where the data to build AI comes from."

MIT Technology Review (December 18,).

Helpman, Elhanan (1998). General purpose technologies and economic growth. Cambridge, MIT Press.

Henrich, Joseph Patrick (2016). The secret of our success : how culture is driving human evolution, domesticating our species, and making us smarter. Princeton, Princeton University Press.

Henrich, Joseph, et al. (2010). "The weirdest people in the world?" Behavioral and Brain Sciences 33(2-3): 61-83.

Horton, John J. (2023) "Large Language Models as Simulated Economic Agents: What Can We Learn from Homo Silicus?", arXiv:2301.07543 DOI: 10.48550/arXiv.2301.07543.

Hu, Edward J., et al. (2021). "LoRA: Low-Rank Adaptation of Large Language Models." arXiv.

Jain, Samyak, et al. (2023). "Mechanistically analyzing the effects of fine-tuning on procedurally defined tasks." arXiv: arXiv:2311.12786.

Jowsey, Tanisha, et al. (2025). "We Reject the Use of Generative Artificial Intelligence for Reflexive Qualitative Research." Qualitative Inquiry: 10778004251401851.

Kalman, David Zvi (2024). "Artificial Intelligence and Jewish Thought". The Cambridge Companion to Religion and Artificial Intelligence. Beth Singer and Fraser Watts. Cambridge, Cambridge University Press: 69-87.

Kaplan, Jared, et al. (2020). "Scaling Laws for Neural Language Models." arXiv (1/23).

Khandelwal, Urvashi, et al. (2019) "Generalization through Memorization: Nearest Neighbor Language Models." arXiv, arXiv:1911.00172 DOI: 10.48550/arXiv.1911.00172.

Kohlberg, Lawrence (1981). The Philosophy of Moral Development: Moral Stages and the Idea of Justice. New York, HarperCollins.

Kuhn, Thomas (1962). The Structure of Scientific Revolutions. Chicago, University of Chicago Press.

Lasswell, Harold (1948). "The Structure and Function of Communications in Society". The Communication of Ideas. Lyman Bryson. New York, Harper: 37-51.

Latour, Bruno (1987). Science in Action: How to Follow Scientists and Engineers Through Society. Milton Keynes, Open University Press.

Lazer, David M. J., et al. (2020). "Computational social science: Obstacles and opportunities." Science 369(6507): 1060-1062.

LeCun, Yann, et al. (2015). "Deep Learning." Nature 521: 436-444.

Lester, Brian, et al. (2021) "The Power of Scale for Parameter-Efficient Prompt Tuning." arXiv, arXiv:2104.08691 DOI: 10.48550/arXiv.2104.08691.

Linegar, et al. (2023). "Large language models and political science." Frontiers in Political Science.

Liu, Yang, et al. (2024) "Datasets for Large Language Models: A Comprehensive Survey." arXiv:2402.18041

Liu, Yang, et al. (2025) "On the Alignment of Large Language Models with Global Human Opinion." arXiv:2509.01418

Magee, Liam, et al. (2023). "Structured like a language model: Analysing AI as an automated subject." Big Data & Society 10(2): 20539517231210273.

Manning, Christopher D., et al. (2020). "Emergent linguistic structure in artificial neural networks trained by self-supervision." Proceedings of the National Academy of Sciences 117(48): 30046-30054.

McCarthy, John (1959). "Programs with common sense." Proceedings of the Teddington Conference on the Mechanization of Thought Processes.

McCarthy, John, et al. (1955). "A Proposal for the Dartmouth Summer Research Project on Artificial Intelligence " Rockefeller Foundation.

Michel, Jean-Baptiste, et al. (2011). "Quantitative Analysis of Culture Using Millions of Digitized Books." Science 331(6014): 176-182.

Mims, Christopher (2025) "We now know how AI "thinks"—and it's barely thinking at all."

Minsky, Marvin and Seymour Papert (1969). Perceptrons an introduction to computational geometry. Cambridge, MIT Press.

Mitchell, Margaret, et al. (2018) "Model Cards for Model Reporting." arXiv, arXiv:1810.03993 DOI: 10.48550/arXiv.1810.03993.

Morrone, Megan (2024). "Google pauses AI image generation after diversity controversies." Axios (Feb 23).

Murdock, Graham P. (1971). Outline of Cultural Materials. New Haven, Human Relations Area Files Press.

Murdock, Graham P. and Douglas R. White (1969) "Standard cross-cultural sample." Ethnology, 329–369.

Neuman, W. Russell, et al. (2025). ""Amazing, They All Lean Left" -- Analyzing the Political Temperaments of Current LLMs." arXiv: arXiv:2507.08027.

Neuman, W. Russell, et al. (2025). "Analyzing the Ethical Logic of Eight Large Language Models." arXiv 2501.08951

Niu, Qian, et al. (2024) "Large Language Models and Cognitive Science: A Comprehensive Review of Similarities, Differences, and Challenges." arXiv:2409.02387

Olah, Chris (2024). "Mapping the Mind of a Large Language Model." Anthropic.ai.

Ouyang, Long, et al. (2022). "Training language models to follow instructions with human feedback." arXiv: arXiv:2203.02155.

Park, Joon Sung, et al. (2024) "Generative Agent Simulations of 1,000 People." arXiv:2411.10109 DOI: 10.48550/arXiv.2411.10109.

Penedo, Guilherme, et al. (2023). "The RefinedWeb Dataset for Falcon LLM: Outperforming Curated Corpora with Web Data, and Web Data Only." arXiv.

Piaget, Jean (1954). The construction of reality in the child. New York, Basic Books.

Piantadosi, Steven T. and Felix Hill (2022). "Meaning without reference in large language models." arXiv: arXiv:2208.02957.

Prothero, Stephen R (2020). Religion matters: an introduction to the world's religions. New York, Norton

Qu, Yao and Jue Wang (2024). "Performance and biases of Large Language Models in public opinion simulation." Humanities and Social Sciences Communications 11(1): 1095.

Resnik, Philip (2025). "Large Language Models Are Biased Because They Are Large Language Models." Computational Linguistics 51(3): 885-906.

Rossi, L. (2024). "The Problems of LLM-Generated Data in Social Science." Sociologica.

Rumelhart, David E., et al. (1986). "Learning representations by back-propagating errors." Nature 323(6088): 533-536.

Rus, Daniela and Gregory Mone (2024). The mind's mirror : risk and reward in the age of AI. New York, NY, W. W. Norton & Company.

Russell, Stuart J. (2019). Human compatible: artificial intelligence and the problem of control. New York NY, Viking.

Salganik, Matthew J. (2018). Bit by bit: social research in the digital age. Princeton, Princeton University Press.

Schroeder, S., et al. (2025). "Large Language Models Do Not Simulate Human Psychology." arXiv arXiv:2508.06950.

Searle, John R. (1980). "Minds, brains, and programs." Behavioral and Brain Sciences 3(3): 417–457.

Senthilkumar, Pranav, et al. (2024). "Fine-Tuning Language Models for Ethical Ambiguity: A Comparative Study of Alignment with Human Responses." arXiv: arXiv:2410.07826.

Shah, Dhavan V., et al., Eds. (2015). Toward Computational Social Science: Exploiting Big Data in the Digital Age. Philadelphia PA, The ANNALS of the American Academy of Political and Social Science.

Shah, Manan, et al. (2026) "Talking to Base Models." New York University

Shojaee, Parshin, et al. (2025). "The Illusion of Thinking: Understanding the Strengths and Limitations of Reasoning Models via the Lens of Problem Complexity." Apple.

Slattery, Peter, et al. (2024) "The AI Risk Repository: A Comprehensive Meta-Review, Database, and Taxonomy of Risks from Artificial Intelligence." arXiv.

Solaiman, Irene and Christy Dennison (2021). "Process for Adapting Language Models to Society (PALMS) with Values-Targeted Datasets." arXiv: arXiv:2106.10328.

Spelke, Elizabeth S. and Katherine D. Kinzler (2007). "Core knowledge." Developmental Science 10(1): 89-96.

Stephens-Davidowitz, Seth (2017). Everybody lies: big data, new data, and what the Internet can tell us about who we really are. New York, Dey Street Books.

Stone, Philip J., et al. (1966). The General Inquirer. Cambridge, MA, MIT Press.

Sutton, Rich (2019). "The Bitter Lesson." Incomplete Ideas (March 13).

Swidler, Ann (1986). "Culture in Action: Symbols and Strategies." American Sociological Review 51(April): 273-286.

Takemoto, Kazuhiro (2024). "The moral machine experiment on large language models." Royal Society Open Science 11(2).

Talmor, Alon, et al. (2019). "oLMpics -- On what Language Model Pre-training Captures." arXiv: arXiv:1912.13283.

Thomson, Judith Jarvis (1976). "The Trolley Problem." Yale Law Journal 94(6): 1395-1415.

Turing, Alan M. (1950). "Computing Machinery and Intelligence." Mind 49: 433-460.

Varnum, Michael E. W., et al. (2024). "Large Language Models based on historical text could offer informative tools for behavioral science." Proceedings of the National Academy of Sciences 121(42): e2407639121.

Wang, A., et al. (2024). "Large Language Models That Replace Human Participants Can Harmfully Misportray and Flatten Identity Groups." arXiv arXiv:2402.01908.

Wiggins, Dion (2025) "All LLMs Now Perform About the Same. Right?" AI Analysis from the Field.

Windsor, Leah Cathryn (2021). "Advancing Interdisciplinary Work in Computational Communication Science." Political Communication 38(1-2): 182-191.

Yax, Nicolas, et al. (2024). "Studying and improving reasoning in humans and machines." Communications Psychology 2(1): 51.

Yin, Zi, et al. (2023). "Alignment is not sufficient to prevent large language models from generating harmful information: A psychoanalytic perspective." arXiv: arXiv:2311.08487.

Yudkowsky, Eliezer (2023). "Pausing AI Developments Isn't Enough. We Need to Shut it All Down." Time (March 29).

Appendix A Methodological Innovations in Tabular Form

This table summarizes the prospects of methodological innovations posited above in tabular form. Given the early stages of this research the elements of the table are unavoidably speculative and incomplete but map out several promising directions for further inquiry.

| Method | Classical Analogue | What Is Varied | What Is Held Constant | Primary Inferential Target | Valid Inferences | Limited Inferences |
|---|---|---|---|---|---|---|
| **Computational Experiments** | Laboratory experiments (psychology, behavioral economics) | Prompt framing, role assignment, payoff structure, authority cues, moral constraints | Model architecture, weights, task structure | Sensitivity of discursive reasoning to contextual perturbation | Framing effects, normative tradeoffs, conditional reasoning patterns | Individual cognition, psychological competence, causal mechanisms in humans |
| **Synthetic Personas** | Survey research, vignette studies | Persona constraints (role, identity, social position), question wording | Model, response format, sampling procedure | Distribution of culturally plausible responses | Discursive priors, narrative repertoires, normative range | Population prevalence, demographic attitudes, opinion frequencies |
| **Computational Comparative-Historical Analysis** | Comparative historical sociology, diachronic text analysis | Training corpus (time period, language, tradition), adapters, retrieval corpora | Model architecture, probing prompts | Historically contingent discursive structure | Relative differences in values, reasoning styles, moral emphasis across periods | Actual historical belief prevalence, lived experience, causal history |
| **Computational Cognitive Neuroscience** | Cognitive neuroscience, representational analysis | Internal activations, features, layers, attention patterns | Training data, task input | Structure of learned representations | Identification of stable semantic or normative features; representational organization | Claims about consciousness, understanding, or neural equivalence to humans |
| **Computational Ablation Studies** | Lesion studies, institutional counterfactuals | Removal or suppression of data sources, features, alignment layers, adapters | Base model architecture | Symbolic dependencies within learned discourse | Which cultural inputs or representational pathways support specific forms of reasoning | Origins of human behavior; necessity claims about real-world cognition |

Appendix B: Validity Issues in LLM-Based Cultural Analysis

This appendix provides a narrative discussion of the principal validity issues associated with using large language models (LLMs) as instruments for sociological and cultural analysis. It reviews key epistemic risks and interpretive constraints, situating them within established debates in cultural analysis, and computational social science. The goal is to clarify the scope conditions under which claims derived from LLM-based cultural condensates are most useful in analysis.

Construct Validity: What Is Being Observed
The primary construct examined in this study is not individual cognition, motivation, or behavior, but the structure of publicly articulated symbolic discourse. Following long-standing traditions in sociology and anthropology, culture is treated here as a system of shared meanings, narratives, and justificatory repertoires rather than as a direct reflection of private mental states (Berger & Luckmann 1966; Swidler 1986). Large language models are understood as cultural condensates: compressed statistical representations of dominant discursive patterns present in large-scale textual corpora. Treating LLM outputs as symbolic structure rather than cognition avoids the category error of anthropomorphizing models or attributing intentions, beliefs, or understanding where none exists (Bender et al. 2021; Searle 1980).

Data Validity and Representational Skew
The training data underlying contemporary LLMs consist largely of text scraped from the public internet, including books, news, forums, and social media. While unprecedented in scale, this corpus is neither representative of humanity as such nor evenly distributed across cultures, languages, or social positions (Liu et al. 2024). Research on digital trace data has repeatedly shown that online discourse overrepresents populations that are Western, educated, industrialized, rich, and democratic (Henrich et al. 2010), as well as individuals who are disproportionately active, expressive, or institutionally positioned (Hargittai 2008).

As a result, LLM-based cultural condensates should be interpreted as structured reflections of a networked public sphere rather than as neutral summaries of human culture. Importantly, this limitation is not merely a source of bias but also an analytic feature. Which voices dominate, which narratives recur, and which perspectives are absent are themselves sociologically meaningful, revealing the power-weighted and institutionally mediated character of contemporary cultural production (Bourdieu 1991).

Procedural Robustness and Replicability
Because LLM outputs are sensitive to prompting, sampling parameters, and aggregation procedures, procedural robustness is a central validity concern. Small changes in prompt wording, role specification, or response constraints can lead to measurable differences in generated output. Accordingly, prompts must be fully documented, and analytic claims should be robust to reasonable perturbations in phrasing or context (Grimmer and Stewart 2013). Replicability in this context does not require identical textual outputs, but convergence at the level of thematic structure or relative emphasis. Consistency across repeated sampling, across models, or across analytic codings strengthens confidence that observed patterns reflect stable features of the condensate rather than stochastic variation or prompt artifacts.

External Validation and Theoretical Triangulation
Validation of LLM-based cultural analysis relies on triangulation rather than direct correspondence with human-subject data. Convergence with established findings in sociology, anthropology, moral psychology, and cultural theory provides contextual support, while systematic divergences are treated as diagnostic signals rather than failures (Salganik 2017). Importantly, no claim in this study rests solely on model output without grounding in prior theoretical or empirical work.

This approach aligns with long-standing practices in interpretive sociology, where texts, narratives, and symbolic systems are analyzed as data in their own right. As with other observational sciences, credibility emerges cumulatively through comparative analysis, replication across contexts, and the disciplined articulation of scope conditions (Denzin 1978).

Scope Conditions and Limits of Inference
Finally, the scope of inference in this study is deliberately narrow. No causal claims about human behavior are advanced, nor are population-level prevalence estimates inferred from model output. The condensates examined here reflect contemporary, digitized discourse and therefore carry temporal and technological specificity. Generalization beyond symbolic expression—to embodied practice, affective experience, or institutional outcomes—requires independent empirical investigation.

Taken together, these validity considerations underscore that LLM-based cultural analysis is best understood as a complementary observational method. Its contribution lies in revealing large-scale patterns of self-description, moral reasoning, and cultural framing that can guide, contextualize, and refine human-centered research, rather than replace it.

Appendix C

Worked Example: Six Llamas — Comparative Religious Ethics Through LoRA-Adapted Language Models

Research Question and Theoretical Motivation

This exemplary computational exercise is drawn from Coleman et al. (2026). It posits the question - can differentially fine-tuned large language models function as useful instruments for comparative cultural analysis? Specifically, if a single base model is adapted on corpora drawn from distinct religious traditions, do the resulting models exhibit systematically different patterns of ethical reasoning that correspond to the moral logics described in the comparative religion and ethics literatures?

The Six Llamas study operationalizes the core claim of the Third Ambition: that large-scale language models encode recoverable cultural structure derived from symbolic human behavior. By constructing multiple "condensates" that share an identical architecture and differ only in the cultural material used for adaptation, the study tests whether ethical reasoning varies in tradition-consistent, interpretable, and falsifiable ways (Fasching et al. 2011; Gudorf 2013; Prothero 2020).

Comparative Condensate Design

Six models were constructed: one unmodified base model and five LoRA-adapted variants, each trained on the canonical texts of a single religious tradition—Christianity, Judaism, Islam, Hinduism, and Buddhism. All models were probed with an identical battery of ethical prompts spanning moral dilemmas, game-theoretic scenarios, public-policy judgments, and moral-psychological self-assessment.

Because LoRA adaptation freezes the base parameters and trains only low-rank adapter matrices, all six models share an identical representational core. Differences in generative output can therefore be attributed to differential cultural exposure rather than architecture, prompt framing, or inference conditions. Methodologically, this mirrors modular comparison strategies in comparative ethics and cultural psychology, while offering stronger control than prompt-only conditioning.

Model Construction and Training

Base model.

Meta-Llama-3.1-8B-Instruct served as both the control condition and the foundation for all adaptations. Models were loaded in 4-bit quantization using Unsloth for efficient training and inference.

Religious corpora.

Each LoRA variant was fine-tuned on a corpus drawn from the Internet Sacred Text Archive, emphasizing canonical and historically influential texts:

| Tradition | Tokens | Corpus Contents |
|---|---|---|
| Christianity | 51,618,717 | Bible translations, apocrypha, early Christian/Gnostic literatures, Reformation texts, modern theology. |
| Hinduism | 18,824,296 | Vedas, Upanishads, Puranas, Mahabharata, Ramayana, Bhagavad Gita, modern commentaries. |
| Judaism | 10,655,984 | Talmud, Midrash, Kabbalah, Babylonian Torah, Haggadah, modern ethical commentary. |
| Buddhism | 5,995,563 | Jataka, Dhammapada, Sutta Nipata, Vinaya texts, Mahayana texts, modern commentaries. |
| Islam | 5,061,927 | Qur'an, Hadith (25,000 sayings), Sufi texts, modern commentaries. |

Training used a standard causal language-modeling objective with identical hyperparameters across traditions. An 80/20 split enabled validation monitoring. This design ensures that variation in outputs reflects differences in ethical tradition, not training regime.

Ethical Prompt Battery

All models received the same 6 prompts, grouped into two domains:

Classic moral dilemmas (Trolley, Footbridge, Heinz, Lifeboat), long used to differentiate deontological and consequentialist reasoning (Foot 1967; Thomson 1976; Kohlberg 1981).

Game-theoretic scenarios (Dictator/Ultimatum Game; Prisoner's Dilemma), probing fairness, generosity, and cooperation.

Each prompt required a forced choice, a justification, and a confidence rating.

Sampling Protocol

All prompts were run under identical generation settings (temperature = 0.5; max = 600 tokens). Each prompt was presented in isolation, with conversation state reset between runs, allowing each response to be treated as an independent observation of the model's ethical condensate.

The unmodified Llama model functions as a baseline approximation of the dominant moral register of contemporary digital text—predominantly Western, English-language, and secular—against which tradition-specific shifts can be interpreted.

Results

Moral Dilemma Responses

| Dilemma | Measure | Llama LLM | Christian | Jewish | Islam | Hindu | Buddhist |
|---|---|---|---|---|---|---|---|
| Trolley | Action | Pull | Decline | Pull | Pull | Pull | Pull |
| Footbridge | Action | Not push | Decline | Push | Push | Self/Fat Man | Not push |
| Heinz | Action | Steal | Not steal | Decline | Steal | Steal | Not steal |
| Lifeboat | Leave behind | Self | Self | Artist | Self | Elderly woman | Elderly woman |
| Dictator | $ offered | $40 | $50 | $100 | $100 | $25 | $50 |
| Prisoners | Testify? | Testify | Not | Not | Testify | Not | Not |

Several structured patterns emerge. The standard Trolley Problem produces near-consensus, while the Footbridge variant sharply divides the models, consistent with long-standing distinctions between indirect and direct harm in both philosophical and religious ethics. The Heinz Dilemma and Dictator Game show particularly wide dispersion, revealing divergent norms concerning property, obligation, and distributive justice.

Importantly, divergence appears where comparative religious ethics predicts disagreement, not randomly. The Jewish model's reluctance to issue a verdict in Heinz echoes rabbinic traditions emphasizing legal interpretation over abstract moralizing (Kalman 2024). The Buddhist model's refusal to endorse killing or theft despite compassionate intent reflects the tension between karuṇā (compassion) and śīla (precepts) documented in Buddhist ethics (Gould & Nishimura 2024). Islamic and Jewish models' generosity in the Dictator Game aligns with traditions emphasizing obligatory charity (zakāt, tzedakah) rather than discretionary benevolence (Chaudhary 2024; Prothero 2020).

Convergence and Divergence

High-convergence domains include the standard Trolley Problem and the Prisoner's Dilemma, suggesting either broad cross-cultural consensus or distance from explicit religious teaching. High-divergence domains—Footbridge, Heinz, abortion, taxation, redistribution—map closely onto areas of genuine doctrinal disagreement in comparative ethics (Fasching et al. 2011; Gudorf 2013). This selective divergence provides evidence that LoRA adaptation captures structured moral differences, not noise.

Interpretation

First, LoRA-adapted models express recognizable moral voices. Christian outputs emphasize rule-based constraint and moral absolutes; Jewish outputs emphasize deliberation and legal framing; Islamic outputs integrate divine authority with welfare maximization; Hindu outputs balance duty (dharma) with pragmatic outcomes; Buddhist outputs constrain compassion with non-harm precepts. These patterns align closely with the comparative ethics literature (Fasching et al. 2011; Prothero 2020; Gudorf 2013).

Second, the base model's secular, consequentialist profile is itself informative, reflecting the moral register of its pretraining corpus rather than neutrality. Its proximity to Buddhism on some dimensions underscores that "secular" does not imply absence of moral structure but the presence of a culturally specific one.

Third, the models preserve internal tensions documented within traditions. The Buddhist case—compassion constrained by absolute prohibitions—is especially illustrative, suggesting that adaptation captures structural features of moral reasoning, not just surface positions.

Falsifiability Conditions

The approach would be undermined by:

> (1) absence of systematic differentiation;
>
> (2) divergence unrelated to known ethical traditions;
>
> (3) instability across repeated runs; or
>
> (4) failure to triangulate with independent religious-ethics scholarship or empirical data.

Conversely, credibility increases as tradition-specific divergence aligns with established ethical theory and remains stable under perturbation.

Limitations and Next Steps

Key limitations include corpus-size asymmetry, reliance on English translations, single-run sampling, and qualitative dimensional coding. Planned extensions include corpus-controlled comparisons, multi-temperature sampling, systematic coding with inter-rater reliability, comparison to non-instruct base models, and interpretability analysis of internal representations.

Summary

As a worked example, Six Llamas demonstrates that comparative condensate analysis is operationally feasible, theoretically grounded, and empirically falsifiable. By holding architecture constant and varying only cultural training data, the study shows that LLMs can encode and express tradition-consistent ethical reasoning, supporting the Third Ambition's claim that language models can serve as instruments for comparative cultural science.